# Hybrid Systems for Knowledge Representation in Artificial Intelligence

Rajeswari P. V. N.
Assoc. Professor, Dept of Comp. Sc. & Engg.
Visvodaya Technical Academy
Kavali, AP, India

Dr. T. V. Prasad
Dean of Computing Sciences
Visvodaya Technical Academy
Kavali, AP, India

*Abstract*— There are few knowledge representation (KR) techniques available for efficiently representing knowledge. However, with the increase in complexity, better methods are needed. Some researchers came up with hybrid mechanisms by combining two or more methods. In an effort to construct an intelligent computer system, a primary consideration is to represent large amounts of knowledge in a way that allows effective use and efficiently organizing information to facilitate making the recommended inferences. There are merits and demerits of combinations, and standardized method of KR is needed. In this paper, various hybrid schemes of KR were explored at length and details presented.

I. INTRODUCTION

An expert (knowledge based) system is a problem solving and decision making system based on knowledge of its task and logical rules or procedures for using knowledge. Both the knowledge and the logic is obtained from the experience of a specialist in the area (Business Expert). An expert system emulates the interaction a user might have with a human expert to solve a problem. The end user provides input by selecting one or more answers from a list or by entering data. The program will ask questions until it has reached a logical conclusion.

*A. Knowledge Engineering*

As described in [1], KR is the process of designing an expert system. It consists of three stages:

- Knowledge acquisition: The process of obtaining the knowledge from experts (by interviewing and/or observing human experts, reading specific books, etc).
- Knowledge representation: Selecting the most appropriate structures to represent the knowledge (lists, sets, scripts, decision trees, object-attribute value triplets, etc).
- Knowledge validation: Testing that the knowledge of ES is correct and complete.

*B. Types of Knowledge*

- Declarative: It describes what is known about a problem. This includes simple statements which are asserted to be either true or false.
- Procedural: Describes how a problem is solved. It contains rules, strategies, agendas and procedures.
- Heuristic: It describes a rule-of-thumb that helps to guides the reasoning process.
- Meta knowledge: Describes knowledge about knowledge for improve the efficiency of problem solving.
- Structural knowledge: It describes about knowledge structures. It contains rule sets, concept relationships and concept to object relationships. [2]
- Factual Knowledge: It is verifiable through experiments and formal methods,
- Tacit knowledge: It is implicit, unconscious knowledge that can be difficult to express in words or other representations form.
- Priori/Prior knowledge: It is independent of the experience or empirical evidence e.g. "everybody born before 1990 is older than 15 years"
- Posteriori/Posterior knowledge: dependent of experience or empirical evidence, as "X was born in 1990".

*C. The Knowledge Representation*

It is an area of AI research which is aimed at representing knowledge in symbols to facilitate inference from those knowledge elements, creating new elements of knowledge, whereas knowledge (is a progression from data to information, from information to knowledge and knowledge to wisdom) and representation ( is a combination of syntax, semantics and reasoning) [3].

There are two basic components of KR i.e. reasoning and inference. In cognitive science it is concerned with how people store and process information and in AI the objective is to store knowledge so that programs can process it. [4]

*D. Knowledge Representation Issues*

The following are the issues to be considered regarding the knowledge representation

- Grain Size – Resolution Detail
- Scope
- Modularity
- Understandability
- Explicit Vs. Implicit Knowledge
- Procedural Vs. Declarative knowledge.

## II. KNOWLEDGE REPRESENTATION TECHNIQUES

Many of the problems in AI require extensive knowledge about the world. Objects, properties, categories and relations between objects, situations, events, states and time, causes and effects are the things that AI needs to represents. KR provides the way to represent all the above defined things [5]. KR techniques are divided in to two major categories that are declarative representation and procedural representation. The declarative representation techniques are used to represents objects, facts, relations. Whereas the procedural representation are used to represent the action performed by the objects. Some of the techniques for knowledge representation are

- Bayesian Network
- Facts and Production Rules
- Semantic nets
- Conceptual Dependency
- CYC
- Frames
- Scripts
- Neural Networks

Hybrid Representation

## III. HYBRID SYSTEMS

A hybrid KR system is an implementation of a hybrid KR formalism consisting of two or more different sub formalisms. These sub formalism should be integrated through (i) a representational theory, which explains what knowledge is to be-represented by what formalism, and (ii) a common semantics for the overall formalism, explaining in a semantic sound manner the relationship between expressions of different sub formalisms.[6] The generalized architecture for a hybrid system is given in Fig 1.

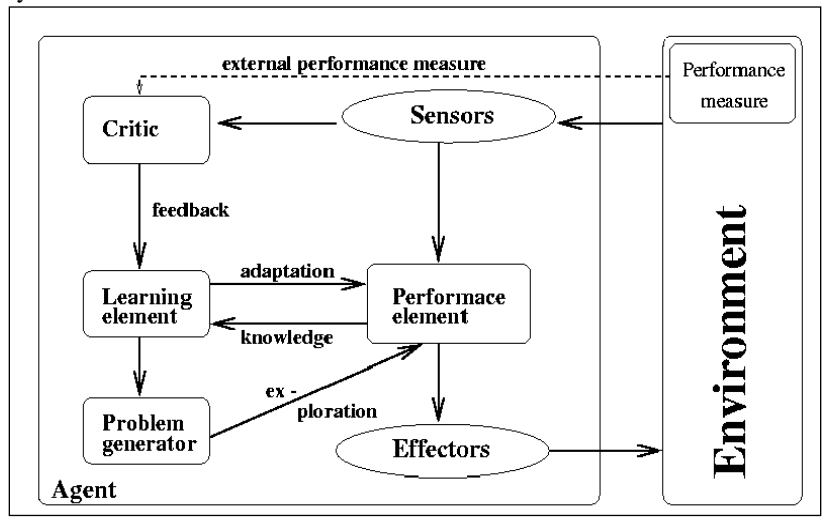

Fig 1 Generalized architecture of Hybrid system

In general these systems consist of two different kinds of knowledge: The terminological knowledge, consisting of a set of concepts and roles defining a terminology, and the assertional knowledge, consisting of some logical formalism suited to represent general assertions.

### A. KRYPTON

The system consists of two modules: the Terminological Box and the Assertional Box. The terminological box, or module, is based on the KL-ONE language -a representation system based on semantic networks and frames [7]. The KRYPTON has been developed mainly from the work of KL-ONE. The difficulties in representing assertional knowledge using KL-ONE gives the idea of the integration of a theorem-prover and a KL-ONE-like language into a hybrid system. It is basically like a "tell-ask" module. All interactions between a user and a KRYPTON knowledge base are mediated by TELL and ASK operations shown in Fig 2.

The most important feature introduced by KRYPTON is the notion of a Functional Approach to knowledge representation [8]: KRYPTON is provided with a clear, implementation independent, description of what services are provided to the user. This Knowledge Level [9] description is presented in the form of a formal definition of the syntax and semantics of the languages provided by the two modules along with the interaction between these two modules.

The set of primitives of the KRYPTON language vary from one presentation to another presentation of the language. In the complete form, the terminological box includes primitives for: Concept conjunction, value and number restriction on concepts, primitive sub-concept, concept decomposition, role differentiation, role chain, primitive subrole and role decomposition. And the assertional box provides a complete first-order logic language including the usual operators: Not, and, or, exists and for all.

### B. KANDOR

The basic units of KANDOR are individuals and frames. Knowledge model for KANDOR is given in Fig 3. Individuals are associated to objects in the real world and frames are associated to sets of these individuals. These units are manipulated through the standard representational structures of frames, slots, restrictions, and slot fillers common to most frame-based systems. Each slot maps individuals into sets of values, called slot fillers, Elements of these sets can be other individuals, strings, or numbers.

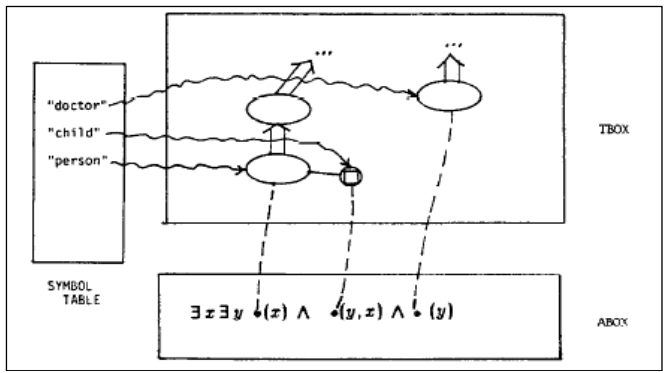

Fig. 2 Overview of KRYPTON

Frames in KANDOR have no assertion import; they look simply as descriptions of some set of individuals. There are two types of frames: Primitive and defined. To be an instance of a primitive frame, an individual must be explicitly specified as an instance of the frame when it is created.

To be an instance of a defined frame an individual must satisfy the conditions associated to the frame definition. There two types of conditions: Super-frames and restrictions. A super-frame is just another frame, and a restriction is a condition on a set of slots fillers for some slot. An individual satisfies the restriction if its slots fillers for that slot satisfy the condition.

KANDOR provides two main operations that require inferences to be made: Given an individual and a frame, determine whether the individual is an instance of the frame, arid, given two frames, it determines whether one frame is subset of another frame.

KANDOR has been used as the knowledge representation component of ARGON [10], which is an interactive information retrieval system which is designed to be used by non experts for retrieval purpose over a large, heterogeneous knowledge bases, possibly taken from a large number of sources or repositories.

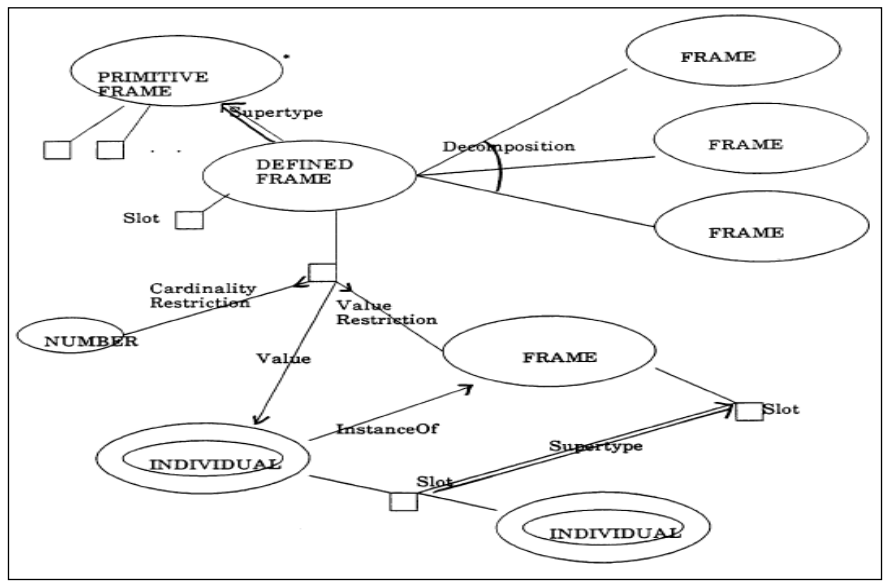

Fig 3 Knowledge model of KANDOR [11]

### C. BACK

The structure of a BACK represents as the same structure of KRYPTON, which contains an terminological box and an assertional box. One main aspect in the BACK implementation is the Balancedness of the formalisms involved. Although the fact that the reasoning in hybrid systems is frequently incomplete (because of efficiency requirements) sometimes leads to situations where one formalism allows to express something which obviously should have some impact on another formalism according to the semantics of the system, the incompleteness of the reasoning precludes this impact to be realized by the system. The formalisms of this type of systems are said to be "unbalanced".

The main criteria taken into account in the development of the BACK system [16] are the following: (i) The sub formalisms of the system should be balanced, (ii) the formalism should permit tractable inference algorithms covering almost all possible inferences, (iii) the assertional box formalism should be able to represent incomplete knowledge in a limited manner, (iv) the system should allow for extending the knowledge base incrementally (retractions are not considered) and (v) the system should reject assertional box entries which are inconsistent. The terminological language of BACK is more powerful than that of KRYPTON.

### D. KL-TWO

The KL-TWO system is composed by two sub formalisms: PENNI, a modified version of the RUP (Reasoning Utility Package) system) and NIKL (New implementation of KL-ONE), a terminological reasoner in the KL-ONE [7] tradition. These two formalisms are complementary: PENNI is able to represent propositional assertions without any quantification and NIKL allows the representation of a simple class of universally quantified sentences. These sentences can be applied in PENNI to extend its propositional language with a limited form of quantification. Fig 4 shows the architecture of KL-TWO [12] .

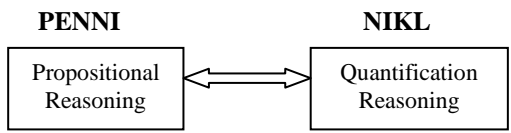

Fig 4 Architecture of KL-TWO

The PENNI formalism consists of a database of propositional assertions, more specifically, a data base of ground sentences of first-order logic without quantifiers. This database permits incremental assertions and retractions. Underlying the deductive mechanism of PENNI is a Truth Maintenance System (TMS) allowing all the useful operations that have been associated with such systems.

And the NIKL terminological reasoner allows the definition of composite concepts and roles through the use of structuring primitives and primitive concepts and roles. The primitives available in NIKL include: Concept conjunction, statement of the minimal number of role fillers, concept value restriction and role differentiation. The inference provided by NIKL is basically the sub assumption relation between concepts. It has been proved recently that the subsumption problem in NIKL is undecidable.

Two forms of hybrid reasoning are performed by the KL-TWO system: The forward reasoning, which is used to classify new assertions according to the concepts already defined in the NIKL knowledge base, and the backward reasoning, used to answer queries. Both mechanisms combine the inferences mechanism of PENNI and NIKL to perform their tasks

*E. CAKE*

The CAKE system was developed as a knowledge representation and reasoning facility for the Programmer's Apprentice project [13] different from the previously presented hybrid systems. CAKE does not present complementary representation formalisms in which different types of knowledge are represented, but it uses its two formalisms to represent the same knowledge.

The two formalisms present in CAKE are: A predicate calculus package which is based on the RUP (Reasoning Utility Package) system and a specialized, semantic network like formalism which is used to represent the structure of programs. This last formalism, called Plan Diagrams or simply Plans, was developed without any special concern about formal semantics but was only designed to fit the requirements of the program representation problem.

The current architecture of CAKE consists of eight layers: The bottom five layers forming the predicate calculus level and the top three layers corresponding to the Plan level. The predicate calculus layers, from bottom to top, and their functions are the following: (i) Truth Maintenance, unit propositional resolution, retraction and explanation, (ii) Equality, uniqueness of terms, congruence closure, (iii) Demons, pattern directed invocation, priority queues, (iv) Algebraic, commutativity, associativity, etc, lattices, Boolean algebras, (v) Types, type inheritance and functionality. The Plan layers are the following: (i) Plan Calculus, data and control flow graphs, abstract data types, (ii) Plan Recognition, flow graph parsing and recognition heuristics, (iii) Plan Synthesis, search of refinement space and synthesis heuristics.

*F. MANTRA [14]*

Developed by J. Calmet, I. A. Tjandra and G. Bittencourt in 1991, it is combination of four different knowledge representation techniques. First order logic, terminological language, semantic networks and Production systems. All algorithm used for inference are decidable because this representation used the four value logic. Mantra is a three layers architecture model. It consist the epistemological level, the logical level, Heuristic level.

Example [1]:- Ex of operation in logic level
1 command::= tell (know1edge base, Fact).
2 ask (knowledge base, Query)
3 to-frames (frame-def)
4 to-met (snet-den)
5 Fact: = to-logic (formu1a)
6 Query: = from logic (formula)
Ex of operation on terminological box
frame - def ::= identifier : c = concept | identifier:
r = relation
Concept::= ( concept ) | concept .

*Advantages:* 1 An intelligent, graphical user interface would help in building knowledge bases. 2 Support procedural knowledge. 3 A graph editor can be used t o visualize, for instance, hierarchies or terminologies that would help the user for representing expert's knowledge.

*Disadvantages:* Less expressive, only applicable for symbolic computation (mathematical model**).**

*G. FRORL*

The acronym for FRORL is Frame-and-Rule Oriented Requirement specification Language [14] which was developed by Jeffrey J. P. Tsai, Thomas Weigert and Hung-Chin Jang in 1992, and this FRORL is based on the concepts of frames and production rules which is mainly designed for software requirement and specification analysis. Six main steps for processing purpose are as follows:

1) Identify subject and themes
2) Define object frames.
3) Define object abstract inheritance relation
4) Define object attributes.
5) Identify activity frames.
6) Define actions and communication

There are two types of frames, i.e., Object frame and Activity frames. Object frames are used to represent the real world entity not limited to physical entity. These frames will act as a data structure. Each activity in FRORL are represented by activity frame to represent the changes in the world. Activity, precondition and action are reserved word not to be used in specification. Language for FRORL consists of Horn clause of predicate logic.

*Advantages:* 1 Modularity. 2 Incremental development. 3 Reusability. 4 Prototyping.

*Disadvantages:* Only limited for building prototype model for software.

### H. Other Hybrid Systems

Other hybrid systems adopt a restricted version of first-order logic in their assertional module. Examples of some latest hybrid systems are

- The LOOM system is a very ambitious project developed at USC (California, USA). It includes a tern classifier, instance matcher, truth maintenance for both TBox and ABox, default reasoning, full-first-order retrieval language, pattern-driven methods, a pattern classifier, and automatic detection of inconsistency. The system can also interface with a rule-based system. Its semantics uses a three-valued semantics that is extended to seven values when defaults are included.
- The QUARK system developed at the University of Hamburg (FRG), includes a Hom clauses interpreter, and a terminological reasoner in the KL-ONE tradition. Its semantics is defined using the four-valued approach. The system is organized around nodes called Denotasional Entities (DE), and the set of facts associated to these DEs, called aggregates. The aggregates correspond to the frame entities in other systems, and are organized into a network.
- The CLASSIC system is a direct descendent of the KANDOR system, and shares all functionalities with this system. The goal of a CLASSIC hybrid system is to extend the expressive power of KANDOR's terminological language while remaining tractable. Along with the functionalities of the KANDOR system, it includes
  (a) a construct to allow equalities between role fillers,
  (b) a set construct to allow one to say a slot is filled by an individual of one of a set of different frames,
  (c) a test-defined construct which allows one to test membership in classes by a user defined test, and
  (d) a limited form of rules which allow one to say that once something is found to be an instance of one concept, then it is an instance of another concept. The system also allows host concepts, such as integers, strings, and all Common Lisp structures.

The Comparison between different hybrid systems [15] is presented in Table 1.

## IV. CONCLUSION

Different KR schemes are used in AI, which differ in terms of semantics, structure and flexibility in level of power of expression. Combination of two or more representation schemes, which is known as Hybrid Systems may be used for making the system more efficient and improving the knowledge representation. Different hybrid systems are discussed with their corresponding architectures and also presented a comparative data in terms of modules (Assertional, Terminology), Formal semantics and Domain of Applications.

AUTHORS PROFILE

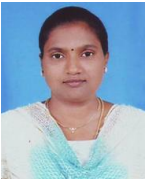

**Ms. P. V. N. Rajeswari** received her Masters degree in Computer Sc. & Engg. from Sam Higgin Bothams University, Allahabad in 2006. She is currently associated with the Dept. of Comp. Sc. & Engg. at Visvodaya Technical Academy, Kavali, AP, India as Associate Professor. She has about 7 years of teaching experience at under graduate and graduate level. Her areas of interest are network security, artificial intelligence, mobile ad-hoc networks, data mining, etc.

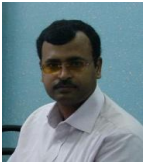

**Dr. T. V. Prasad** has over 17 years of experience in industry and academics. He has received his graduate and master's degree in Computer Science from Nagarjuna University, AP, India. He was with the Bureau of Indian Standards, New Delhi for 11 years as Scientist/Deputy Director. He earned PhD from Jamia Millia Islamia University, New Delhi in the area of computer sciences/ bioinformatics. He has worked as Head of the Department of Computer Science & Engineering, Dean of R&D and Industrial Consultancy and then as Dean of Academic Affairs at Lingaya's University, Faridabad. He is with Visvodaya Technical Academy, Kavali as Dean of Computing Sciences. He has lectured at various international and national forums on subjects related to computing.

Prof. Prasad is a member of IEEE, IAENG, Computer Society of India (CSI), and life member of Indian Society of Remote Sensing (ISRS) and APBioNet. His research interests include bioinformatics, consciousness studies, artificial intelligence (natural language processing, swarm intelligence, robotics, BCI, knowledge representation and retrieval). He has over 73 papers in different journals and conferences, and has authored six books and two chapters.

TABLE 1  COMPARISON OF HYBRID SYSTEMS

|  | **Assertional Module** | **Terminology Module** | **Other Modules** | **Formal Semantics** | **Domain of Application** |
|---|---|---|---|---|---|
| **KRYPTON** | Full first order predicate logic | KL-ONE like | - | Model Theoretic | Natural Language |
| **KANDOR** | Frame like schema | KL-ONE like | - | Model Theoretic | - |
| **BACK** | Object oriented language | KL-ONE like | - | Model Theoretic | Natural Language |
| **KL-1WO** | Variable free predicate logic | KL-ONE like | - | Model Theoretic | - |
| **CAKE** | Truth maintenance system | - | Plan diagrams | Mapping into logic | Programming Language |
| **LOOM** | Truth maintenance system | Variable-free algebra | Production systems | Seven-valued semantics | Natural Language |
| **QUARK** | Horn classes | KL-ONE like | - | Four-valued semantic | Natural Language |
| **CLASSIC** | Frame like schema | KL-ONE like | - | Model Theoretic | Prototype application : Wine Choice |
| **DRL** | Prolog | Many sorted theory | - | Prolog semantics | Logic Programming |
| **KRAPFEN** | Network of propositions | KL-ONE like | Proto type module | Not provided | Natural Language |
| **MANTRA** | Decidable first order logic | KL-ONE like | Semantic net and production systems | Four-valued semantic | Mathematical Knowledge |